%% file: main.tex
\documentclass[sigconf,natbib=false]{acmart}  % ,nonacm
\usepackage[numbers]{natbib}
\usepackage[utf8]{inputenc} % allow utf-8 input
\usepackage[T1]{fontenc}    % use 8-bit T1 fonts
\usepackage{hyperref}       % hyperlinks
\usepackage{cleveref}
\usepackage{url}            % simple URL typesetting
\usepackage{booktabs}       % professional-quality tables
\usepackage{amsfonts}       % blackboard math symbols
\usepackage{amsmath}       % blackboard math symbols
\usepackage{nicefrac}       % compact symbols for 1/2, etc.
\usepackage{microtype}      % microtypography
\usepackage{xcolor}         % colors

\usepackage{subcaption}
\usepackage{graphicx, wrapfig}
\usepackage{caption}
\usepackage{tablefootnote}
\usepackage{xspace}
\usepackage[title,titletoc]{appendix}
\usepackage{multirow}
\usepackage{adjustbox}
\usepackage{bm}
\usepackage[ruled]{algorithm2e} % For algorithms%
\usepackage{placeins}
\usepackage{enumitem}

\crefname{algocf}{alg.}{algs.}
\Crefname{algocf}{Algorithm}{Algorithms}

\copyrightyear{2024}
\acmYear{2024}
\setcopyright{acmlicensed}
\acmConference[DEEM 24]{Workshop on Data Management for End-to-End Machine Learning}{June 9, 2024}{Santiago, AA, Chile}
\acmBooktitle{Workshop on Data Management for End-to-End Machine Learning (DEEM 24), June 9, 2024, Santiago, AA, Chile}
\acmDOI{10.1145/3650203.3663331}
\acmISBN{979-8-4007-0611-0/24/06}

\begin{document}

    \title[Federated Fine-Tuning of LLMs on the Very Edge: The Good, the Bad, the Ugly]{Federated Fine-Tuning of LLMs on the Very Edge: \\ The Good, the Bad, the Ugly}

    \author{Herbert Woisetschl\"ager}
    \email{herbert.woisetschlaeger@tum.de}
    \affiliation{%
      \institution{Technical University of Munich}
      \city{Munich}
      \country{Germany}
    }

    \author{Alexander Erben}
    \email{alex.erben@tum.de}
    \affiliation{%
      \institution{Technical University of Munich}
      \city{Munich}
      \country{Germany}
    }

    \author{Shiqiang Wang}
    \email{wangshiq@us.ibm.com}
    \affiliation{%
      \institution{IBM T.J. Watson Research Center}
      \city{Yorktown Heights}
      \country{United States}
    }

    \author{Ruben Mayer}
    \email{ruben.mayer@uni-bayreuth.de}
    \affiliation{%
      \institution{University of Bayreuth}
      \city{Bayreuth}
      \country{Germany}
    }

    \author{Hans-Arno Jacobsen}
    \email{jacobsen@eecg.toronto.edu}
    \affiliation{%
      \institution{University of Toronto}
      \city{Toronto}
      \country{Canada}
    }

    \begin{abstract}
    With the emergence of AI regulations, such as the EU AI Act, requirements for simple data lineage, enforcement of low data bias, and energy efficiency have become a priority for everyone offering AI services.
    Being pre-trained on versatile and a vast amount of data, large language models and foundation models (FMs) offer a good basis for building high-quality deep learning pipelines.
    Fine-tuning can further improve model performance on a specific downstream task, which requires orders of magnitude less data than pre-training. 
    Often, access to high-quality and low-bias data for model fine-tuning is limited due to technical or regulatory requirements.
    Federated learning (FL), as a distributed and privacy-preserving technique, offers a well-suited approach to significantly expanding data access for model fine-tuning. 
    Yet, this data is often located on the network edge, where energy, computational, and communication resources are significantly more limited than in data centers.
    
    \noindent In our paper, we conduct an end-to-end evaluation for fine-tuning the FLAN-T5 FM family on the network edge. 
    We study energy efficiency potentials throughout FL systems - on clients, in communication, and on the server.
    Our analysis introduces energy efficiency as a real-time metric to assess the computational efficiency of an FL system.
    We show the stark need for further improvements in communication efficiency when working with FMs and demonstrate the importance of adaptive FL optimizers for FM training.
    \end{abstract}

\renewcommand{\shortauthors}{Woisetschl\"ager et al.}

\maketitle

% Chapters
    \section{Introduction}
    \label{sec:intro}
    \input{chapters/01_introduction}

    \section{Background}
    \label{sec:background}
    \input{chapters/02_background}

    \section{Methodology}
    \label{sec:methodology}
    \input{chapters/02_methodology}

    \section{Experimental Setup}
    \label{sec:experimental_setup}
    \input{chapters/03_experimental_setup}

    \vspace{-6pt}
    \section{Results}
    \label{sec:evaluation}
    \input{chapters/04_results}

    \section{Related Work}
    \label{sec:related-work}
    \input{chapters/05_related_work}

    \section{Discussion}
    \label{sec:discussion}
    \input{chapters/06_discussion}

    \vspace{8pt}
    \section{Conclusions}
    \label{sec:conclusion}
    \input{chapters/07_conclusion}
    \newpage

    \section*{Acknowledgements}
    This work is partially funded by the Bavarian Ministry of Economic Affairs, Regional Development and Energy (Grant: DIK0446/01), the German Federal Ministry for Economic Affairs and Climate Action (Grant: 16KN085729), and the German Research Foundation (DFG, Grant: 392214008).
    
    {
        \balance
        \small
        \bibliographystyle{plainnat}
        \bibliography{main}
    }
    \newpage

    \begin{appendices}
        \include{chapters/08_appendix}
    \end{appendices}
    \newpage

\end{document}

%% file: chapters/01_introduction.tex
\begin{figure}
    \includegraphics[width=0.48\textwidth]{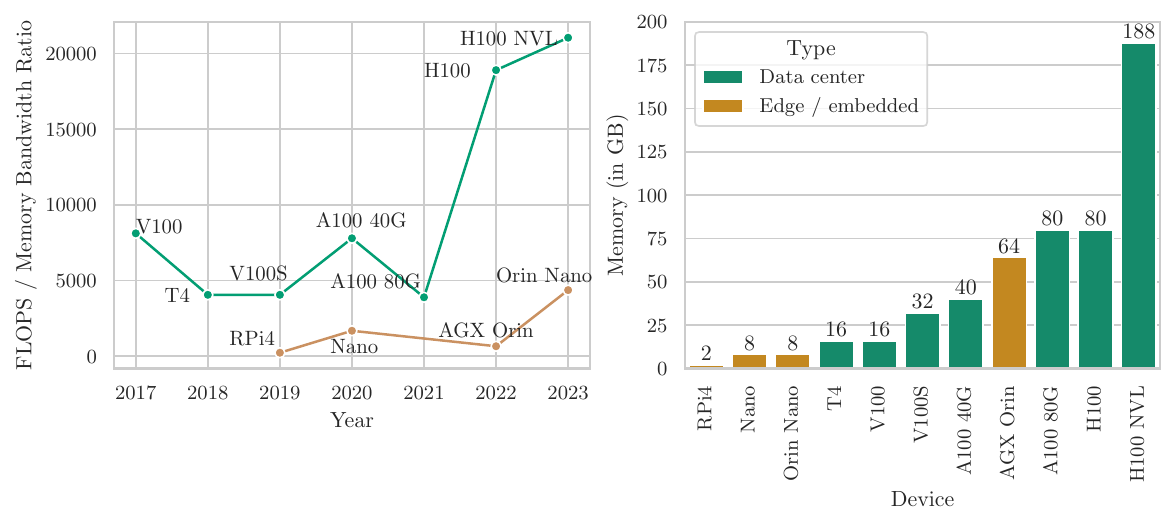}
    \caption{Development of computational power and resource availability of DL accelerators 2017 - 2023 for data centers and embedded systems. Key: RPi4 = Raspberry Pi 4, Nano = NVIDIA Jetson Nano, Orin Nano = NVIDIA Jetson Orin Nano, AGX Orin = NVIDIA Jetson AGX Orin 64 GB.}
    \label{fig:model-flops-hw}
    \vspace{-12pt}
\end{figure}

Large Language Models (LLMs) and Foundation Models (FMs) are omnipresent in academia and practice and fuel new innovations~\cite{bommasani2021opportunities}.
These models have grown significantly with regard to parameter size, as more parameters improve the performance to a certain degree~\cite{hoffmann2022training}. 
In line with the growing computational need for these models, deep learning (DL) hardware accelerators have become increasingly more capable.
Recent developments indicate a generational leap in computational power for data center applications, with the NVIDIA H100 NVL delivering 7.8 TB/s memory bandwidth compared to the previous state-of-the-art A100 80GB GPU that only has 2 TB/s (\Cref{fig:model-flops-hw}).
Due to memory-bandwidth bottlenecked operations taking up to 40\% of the training time~\cite{ivanov2021data}, this improvement may lead to much faster training times for both small and large models.
At the same time, computational capabilities on embedded devices for mobile edge computing are significantly growing, with the NVIDIA Jetson AGX Orin 64GB being the first-of-a-kind DL-accelerated embedded device that provides capabilities for training FMs \cite{Chung2022}. This has never been possible before and enables us to build FL workloads with large transformer models, benefit from scattered data, and bring generative AI closer to users, all the while improving data privacy.

\noindent As this type of device is oftentimes scattered across geographies and entities, federated DL (FL) imposes itself as a well-suited technique for fine-tuning FMs in a distributed and private fashion. 
To our knowledge, the largest models discussed in FL to this point entail FedBert and GPT2~\cite{Tian2022, zhang2023towards}. 
Both models were trained with FL methods on multi-GPU data center nodes. Only a few studies exist on field deployments \cite{Baunsgaard2021}.
As can be seen in \Cref{fig:model-flops-hw}, the computing resources of state-of-the-art embedded hardware like the NVIDIA Jetson AGX Orin are orders of magnitude less than on a modern data center GPU like the H100.
However, if we want to gain access to a broader data basis, we need to foster FL on the edge and bring FMs to embedded devices. 

\noindent At the same time, new regulations like the European Union AI Act impose new limitations and requirements for FL~\cite{woisetschlaeger2024} that need to be met such that applications become practical. 
This entails the need to prioritize energy efficiency. 
For FL, this is an underexplored area as most existing works either perform microbenchmarks for FL clients \cite{fedscale-icml22, beutel2020} or focus on communication cost reduction \cite{feng2021min}.

\noindent The combination of resource limitations and increasing regulatory requirements presents us with a set of challenges:

\begingroup
    \renewcommand\labelenumi{(\theenumi)}
    \begin{enumerate}
        \item \textbf{Comparably low memory bandwidth on embedded devices limits the compute potential of FL applications on the edge}. We currently see a generation leap in data center DL accelerators regarding memory bandwidth, which has increased significantly (up to 7.8 TB/s). Even though the memory size on embedded devices has increased, the memory bandwidth remains comparatively low (up to 0.2 TB/s). This affects key memory-bandwidth bottlenecked operations for the training process, which could lead to severe training time penalties.
        \vspace{6pt}

        \item \textbf{Energy efficiency has become a priority with the introduction of the EU AI Act.} The new regulation requires service providers that offer FL applications to focus on energy-efficient operations. To this point, energy efficiency can be measured by means of the Model-FLOP utilization \cite{Chowdhery2022}, but it requires knowledge of what hardware is being used and how to optimally configure it. In FL systems, this can be impractical as we often do not know client details, and in many cases, clients are likely to participate only once in the training process~\cite{Malinovsky2023}.
        \vspace{6pt}

        \item \textbf{FMs are large in size and are harder to train or fine-tune than small models}. The prevailing benefit of FMs is their ability to cater to many different tasks~\cite{bommasani2021opportunities}. However, their performance on specific downstream tasks can be improved with fine-tuning~\cite{Hu2021}. Yet, the gradients during foundation model training or fine-tuning are at much higher risk of exploding or vanishing than in smaller tasks, as they are frequently discussed in FL~\cite{fedscale-icml22,He2020_fedml,caldas2018}. 
        \vspace{6pt}
        
        \item \textbf{Communication on the edge is significantly more expensive than in data centers}. 
        While network bandwidth in data centers is available at 100~Gbit~\cite{awsp3}, mobile or remote communication over wide area networks is still a difficult challenge to achieve, especially when handling 100M+ parameter DL models. 
        For distributed learning applications where high-bandwidth communication is available, we can use techniques such as ZeRo-offloading~\cite{ren2021zero} and FSDP~\cite{zhao2023pytorch} that utilize a high-bandwidth interconnect for all-reduce communication between nodes to not materialize the full model, optimizer, and gradient state due to limited memory sizes. 
        In FL, this is typically infeasible due to a limited network interconnect.

    \end{enumerate}
\endgroup

\noindent Based on the open challenges to creating efficient edge computing systems capable of training FMs, we formulate our research question: 
How can we efficiently realize FM training and fine-tuning at the network edge?
Which levers can have the largest impact on improving FL system efficiency?

\noindent By exploring this research question, we make four major contributions to bridge the gap between federated foundation model training and energy-aware FL: 

\begingroup
    \renewcommand\labelenumi{(\theenumi)}
    \begin{enumerate}
    
        \item \textbf{We systematically study the computational limitations of state-of-the-art embedded hardware for DL}. Nowadays, most papers in the FL space use data center hardware for their experiments~\cite{He2020_fedml,caldas2018}, while large amounts of data are scattered on the edge and must not be neglected as a field of application. 
        We, therefore, conduct an in-depth micro-benchmark of various transformer models on the latest embedded and datacenter DL accelerators to identify computational bottlenecks.
        \vspace{6pt}

        \item \textbf{We outline the limitations of theoretical metrics such as the Model-FLOP Utilization for FL applications}. 
        As micro-benchmarks require extensive experimentation, practitioners often use metrics, such as the Model-FLOP Utilization (MFU), based on theoretical hardware performance limits to assess the computational efficiency of algorithms~\cite{DESISLAVOV2023100857}.
        Calculating the MFU requires knowledge of hardware specifications on FL clients, which could be infeasible due to privacy considerations. 
        As such, we identify energy efficiency as a readily available alternative to the MFU and outline that the computational limits of embedded AI accelerators appear significantly earlier than the MFU suggests.
        \vspace{6pt}

        \item \textbf{We benchmark four state-of-the-art FL optimizers for FM fine-tuning}. 
        A key to energy-efficient use of FL is the right optimizer choice. 
        We systematically benchmark four state-of-the-art FL optimizers to quantify the energy savings with the right optimizer choice. 
        We find adaptive optimization techniques to converge up to $8\times$ faster than FedAvg with momentum, one of the most widely used FL optimizers \cite{liu2019_momentum}.
        \vspace{6pt}

        \item \textbf{We quantify the \textit{total} cost of communication in FL applications with state-of-the-art FMs}.
        Our study identifies wide-area communication as the primary driver for energy consumption in FL systems, up to 4 orders of magnitude higher than the energy consumed by computing on clients. 

    \end{enumerate}
\endgroup

\noindent This paper is structured as follows. \Cref*{sec:background} will outline relevant background. In \Cref*{sec:methodology}, we present our methodology, and in \Cref*{sec:experimental_setup}, we present our benchmark design, including datasets, DL models, and FL strategies. \Cref*{sec:evaluation} contains experimental evaluations of our benchmark. In \Cref*{sec:related-work}, we present related work. In \Cref*{sec:discussion}, we discuss our results. In \Cref*{sec:conclusion}, we conclude our work.

%% file: chapters/02_background.tex
\subsection{Performance Objectives in Data Center Environments}
One of the most important issues when training in a data center is to maximize throughput by trying to use the hardware to its limit without being blocked by communication.
Communication concerns both local communication, i.e., memory movement, and communication between GPUs and nodes, typically with a high bandwidth interconnect such as NVLink (7.8~TB/s) and Ethernet (100~Gbit)~\cite{awsp3}. 

\noindent Measuring the effectiveness of each GPU in training is possible via Model FLOP Utilization (MFU)~\cite{Chowdhery2022}, which is the ratio of throughput achieved compared to the theoretical throughput of a model and a set of hardware.
Common values for the MFU are between 5 -- 20\% (\Cref{fig:mfu}) because DL models are not defined as a single matrix multiplication that can be perfectly parallelized between tensor cores but as many operations with memory bandwidth bottlenecks such as softmax, residual additions, and activations~\cite{ivanov2021data}.
These operations result in a FLOP usage significantly lower than the theoretical hardware capability, and each model architecture has its own set of operations that slow down throughput.
However, the MFU can be used as a benchmark for how well a model is suited to work on a particular piece of hardware, as it fits the trade-offs between memory bandwidth, memory capacity, and FLOP.
This way, we can compare the MFU for the same model on different hardware and contrast their results.

\subsection{Performance Objectives on the Edge}
In edge computing systems that involve embedded devices, performance considerations differ from those in data center environments as use cases typically vary~\cite{varghese2020}. Yet, to run FL workloads on embedded devices on the edge, we need to unite performance characteristics from data centers and edge computing.

\noindent Running FL workloads on the edge is all about minimizing the time we use a client's hardware and maximizing the throughput. 
Yet, the hardware is often located in remote areas with limited access to power or even on mobile devices with very restrictive battery management~\cite{Ogden2018}. Also, in remote and mobile environments, network bandwidth utilization and total network traffic are critical. Both have a significant impact on communication latency, i.e., how fast we can move model weights between clients and a server. For foundation models, both can become a hurdle, as this kind of model tends to grow beyond several hundreds of millions of parameters in size or, in other words, beyond 1 GB in parameters to transfer over the network. Putting that into perspective with the average available wireless network bandwidth of 50 Mbit/s on mobile devices \cite{errant2021} yields communication times substantially longer than the actual computation time on clients \cite{beutel2020}.

\subsection{Regulatory Requirements with Regard to Energy Efficiency}
As we see regulatory frameworks on Artificial Intelligence emerge and be passed as laws, the first legislation to come into effect by 2024 is the EU AI Act \cite{eu_ai_act}. 
Other countries have declared the EU AI Act as a lighthouse framework; they aim to align their individual frameworks with it \cite{canada_act}.
A priority in the EU AI Act is energy efficiency \cite{woisetschlaeger2024}. 
The objective is to promote sustainable computing practices by holding service providers liable for monitoring energy consumption and subsequently fostering the energy efficiency of an FL system.
As such, it is vital to understand where and how efficiency potentials can be lifted such that the practical applicability of FL is improved.

\begin{algorithm}
\caption{Federated Adam with decoupled weight decay (FedAdamW)}    
\label{alg:fedadamw}
    \input{algorithms/fedadamw}
\end{algorithm}

%% file: algorithms/fedadamw.tex
\SetKwInput{KwData}{Given}
\KwData{set of clients $K \in \mathbb{N}^+$, server rounds $s \in \mathbb{N}^+$, $\beta_1 = 0.9$, $\beta_2 =0.999$, $\tau = 10^{-6}$, $\lambda\in \mathbb{R}$}
\BlankLine
\SetKwInput{KwData}{Initialize}
\KwData{server round $t \leftarrow 0$, initial parameters $x_{t=0} \in \mathbb{R}$, first momentum vector $m_{t=0} \leftarrow 0$, second momentum vector $v_{t=0} \leftarrow 0$, server learning rate $\eta_s \in \mathbb{R}$, client learning rate $\eta_c \in \mathbb{R}$}
\BlankLine

\For{$t$ \KwTo $s$}{

    \BlankLine
    
    \tcp*[h]{server-side}
    
    \emph{Sample subset $k \in K$}
    
    $x^t_{i} = x_t$
    
    \emph{Distribute model weights to clients}

    \BlankLine
    \BlankLine
    \tcp*[h]{client-side (is identical with FedAvg)}
    
    \For{$i \in k$ \textbf{in parallel}}{
        $g^t_i \leftarrow \nabla F_i(x^t_i)$ \tcp*[h]{\footnotesize compute local gradient}
        
        $x^{t+1}_i = x^t_i - \eta_c \cdot g^t_i$ \tcp*[h]{\footnotesize update client model}
        
        \emph{Send model update $x^{t+1}_i$ to server}
    }

    \BlankLine
    \tcp*[h]{server-side optimization}
    
    $x_{t+1} \leftarrow \frac{1}{|k|}\sum_i^{|k|} x^{t+1}_i$  \tcp*[h]{\footnotesize FedAvg}
    
    $g_t \leftarrow x_t - x_{t+1}$ \tcp*[h]{\footnotesize pseudo gradient}
    
    $m_{t+1} \leftarrow \beta_1 \cdot m_t + (1-\beta_1) \cdot g_t$  \tcp*[h]{\footnotesize momentum}
    
    $v_{t+1} \leftarrow \beta_2 \cdot v_t + (1-\beta_2) \cdot g_t^2$  \tcp*[h]{\footnotesize velocity}

    \BlankLine
    \tcp*[h]{FedAdamW}
    
    $x_{t+1} \leftarrow x_{t} - \eta_s \cdot \lambda x_{t}$  \tcp*[h]{\footnotesize decoupled weight decay}
    
    $\hat{m}_{t+1} \leftarrow \frac{m_{t+1}}{1 - \beta_1^t}$ \tcp*[h]{\footnotesize regularized momentum}
    
    $\hat{v}_{t+1} \leftarrow \frac{v_{t+1}}{1 - \beta_2^t}$  \tcp*[h]{\footnotesize regularized velocity}
    
    $x_{t+1} \leftarrow x_{t+1} - \frac{\eta_s \cdot \hat{m}_{t+1}}{\sqrt{\hat{v}_{t+1}} + \tau}$

    $t \leftarrow t + 1$ \tcp*[h]{\footnotesize update FL round}

}
\BlankLine
\KwResult{optimized parameters $x_{t}$}

%% file: chapters/02_methodology.tex
Generally, when transferring LLM fine-tuning to the edge, we also transfer the challenges we currently have in data center environments into systems that suffer from more severe resource limitations. 
While energy efficiency is a specific challenge to edge computing systems, network bandwidth and computational efficiency are frequently discussed topics for DL applications in data centers.
With our hardware-centric study, we aim to provide a comprehensive perspective on energy efficiency levers in FL systems on the edge to foster sustainable computing and, subsequently, legal compliance with the EU AI Act. 
To do so, we organize our methodology along the following four pillars to cover the end-to-end training pipeline.

\subsection{Computational Efficiency}
By studying the behavior of state-of-the-art FL clients when it comes to scaling on-device training with varying model sizes and minibatch sizes, we aim to understand how the training steps (forward, loss calculation, \texttt{opt.step()}, and backward) differ between data center resources and clients deployed on the network edge. 
Maximization of resource utilization is the superior objective for DL and FL applications in data center environments, as this is usually equivalent to a cost-optimal solution~\cite{frey2022}. 
In the HPC domain, MFU is used to calculate the hardware resource utilization based on the number of theoretical hardware FLOP/s. 
By varying the minibatch size, the MFU can also be used to identify computational bottlenecks, i.e., whether we are computationally bound or memory bandwidth limited. 
In our experiments, the theoretical capacity of the NVIDIA A100 is 312~TFLOP (at FP32), while the Jetson AGX Orin 64~GB provides 42.5~TFLOP or 13\% of the A100.
As such, the MFU is well suited for in-depth analysis but requires full knowledge of client hardware specifications and the availability of performance metrics.
However, in FL systems, clients are often heterogeneous regarding their hardware and considered ephemeral, i.e., they are likely to participate in training only once~\cite{Malinovsky2023}. 
% Therefore, we would require an extensive knowledge base about the theoretical computing capabilities of clients that may participate in an FL system to evaluate computational efficiency.

\subsection{Energy Efficiency}
With the imminent legal requirements to focus on energy efficiency, practical FL system design must include energy monitoring, regardless of whether FL clients are deployed in a data center or on the network edge.
Yet, in edge computing, energy efficiency has been a priority for a long time~\cite{shinde2021design,yu2021toward,zheng2021federated}. 
A major benefit of clients on the network edge, such as NVIDIA Jetson AGX Orin, is their hardware design since they are often created as a system-on-a-chip (SoC) and contain hardware-based power measurement units for each component (e.g., CPU, GPU).
In contrast to MFU, which requires detailed hardware knowledge, energy metrics are likely to be readily available and easy to measure across all clients.
We define energy efficiency as the tokens per second ($\mathrm{TPS}$) throughput over the average power draw (W) for a workload,
\begin{equation}
    \label{eq:energy_efficiency}
    \eta_e = \frac{\mathrm{TPS}}{W}\mathrm{.}
\end{equation}

\subsection{Communication Efficiency}
Communication is equally important for federated LLM fine-tuning as computational efficiency. Typically, full models or partial model weights are communicated between client and server \cite{mcmahan2016}. Yet, communication on data center settings is built on top of high-performance networking infrastructure that enables bandwidths of 100~Gbit and more \cite{awsp3}. On the edge, we often find significantly slower network links with 1~Gbit and below. For instance, the global average for communication over 4G LTE wireless is 40~Mbit download and 15~Mbit upload~\cite{errant2021}. 

\noindent We need a reliable metric to quantify communication efficiency that, at the same time, tells us whether it is useful to further scale a FL workload over more clients or not.
Borrowing from the HPC domain, Granularity ($G$) measures the ratio between the time it takes to compute a DL task ($T_{\mathrm{comp}}$) and to communicate the model gradients or weights ($T_{\mathrm{comm}}$)~\cite{10.5555/541880}. It is defined as 
\begin{equation}
    G = \frac{T_{\mathrm{comp}}}{T_{\mathrm{comm}}}\mathrm{.}
\end{equation}

\noindent In our FL scenario, the computation time is the maximum fine-tuning time on a client in each round, and the communication time is the time spent sending the model state, waiting, and receiving the aggregated model state from the server. 
In general, $G \gg 1$ indicates that adding one more client to a system has a positive effect on the total processing speed (higher throughput). 
$G\ll1$ is an indicator for communication times significantly outweighing computation times and, therefore, no positive effect on system throughput.
As such, we use $G$ as the evaluation metric to evaluate the practical utility of federating an FL application.

\noindent In addition to scalability considerations, we evaluate communication costs when deploying FL applications to the network edge.
To do so, we consider two scenarios. 
First, we look at a mobile edge computing scenario where clients are connected via an LTE wireless connection \cite{3gpp_4g_lte}, which exhibits download and upload speeds of 40 and 10 Mbit, respectively \cite{errant2021}.
Second, we consider a scenario where FL clients are operating at the network edge with a wired 1 Gbit connection that is often found in factory settings \cite{Kubiak2022}.
We use the \textit{per-bit communication model} to estimate the total communication cost of our FL pipelines \cite{woisetschlaeger2024,Yousefpour2023, Jalali2014}.
It is important to note that once wireless communication is involved, the energy consumption for communication increases by two orders of magnitude \cite{Jalali2014}. 
A detailed explanation and the exact parameterization of the per-bit communication model are available in \Cref{app:methodology}.

\begin{figure*}[!ht]
    \includegraphics[width=\textwidth]{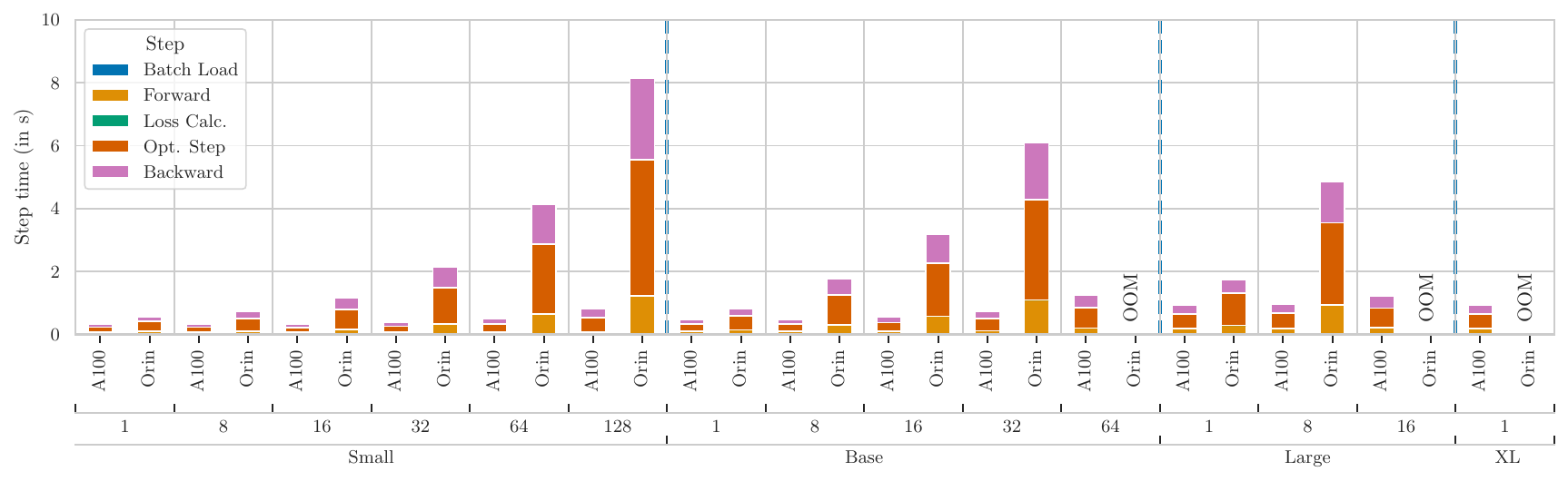}
    \caption{DL training step times across FLAN-T5 transformer models with varying minibatch sizes on the Samsum dataset running on the NVIDIA A100 and Jetson AGX Orin platform. Detailed metrics are available in Appendix \Cref{app:results}.}
    \label{fig:microbench}
\end{figure*}

\subsection{Model Performance}
We use four widely used federated optimizers: (I) Federated Averaging (FedAvg) \cite{mcmahan2016}, (II) FedAvg with Momentum (FedAvgM) \cite{liu2019_momentum}, (III) Federated Adam (FedAdam) \cite{reddi2020}, and (IV) we introduce FedAdam with decoupled weight decay (FedAdamW).
The objective of each optimizer is to minimize the loss of a given neural network, typically done by stochastic gradient descent (SGD).
All four optimizers share SGD as the optimization basis.

\noindent FedAvg is used to control the communication efficiency of an FL application as it allows training over multiple minibatches on a client before communicating a model update. 
With federated SGD, we would have to communicate after each minibatch \cite{mcmahan2016}.
However, as soon as we encounter high change rates in gradients, as is often the case when working with foundation models, we require adaptive control over the model learning rate \cite{kingma2014}.
FedAvgM introduces (first-order) momentum regularization to reduce the impact of early gradient and stabilize training.
Yet, often, this is not enough, as reducing the momentum too much slows down model convergence towards the end of the training; subsequently increasing training costs \cite{kingma2014}.
For this, \citet{reddi2020} introduced FedAdam among other federated optimizers.
Additionally to momentum, FedAdam uses velocity (second-order momentum) to further regularize gradients.
Even though FedAdam may provide faster model convergence, similar to its centralized counterpart Adam, it is challenged by a worse generalization over new data than SGD.

\noindent To tackle this challenge, we implement federated Adam with decoupled weight decay (FedAdamW) based on \citet{Loshchilov2017} (\Cref{alg:fedadamw}). 
A theoretical analysis for decoupling weight decay in adaptive optimization algorithms is provided by \citet{jin2022_fed_da}.
The objective is to use weight decay as an additional optimization technique to penalize large or vanishing gradients. 
To evaluate the effectiveness of each optimization technique, we study the validation loss and the Rouge-1 score. 
In natural language processing research, the Rouge-1 score evaluates unigram overlaps between a model-generated output and a reference text.
Generally, a Rouge-1 score of 50 is considered a strong result.
It is a derivative of the F1-Score known from classification tasks (as often encountered in computer vision research) \cite{lin-2004-rouge}.

%% file: chapters/03_experimental_setup.tex
Our hardware-centric study for FL on the edge focuses on evaluating state-of-the-art DL workloads on embedded devices. As such, we focus on a state-of-the-art FM family. 
% Our code base for further exploration is available online.\footnote{For the review period, the code base is available as an anonymous git repository and will be available on GitHub soon: \url{https://anonymous.4open.science/r/fl_workshop_neurips_2023-04CB}}. Further details are available in Appendix \Cref{app:exp-setup} (esp. FL strategy configuration).

\begin{figure}[!ht]
    \centering
    \includegraphics[width=0.48\textwidth]{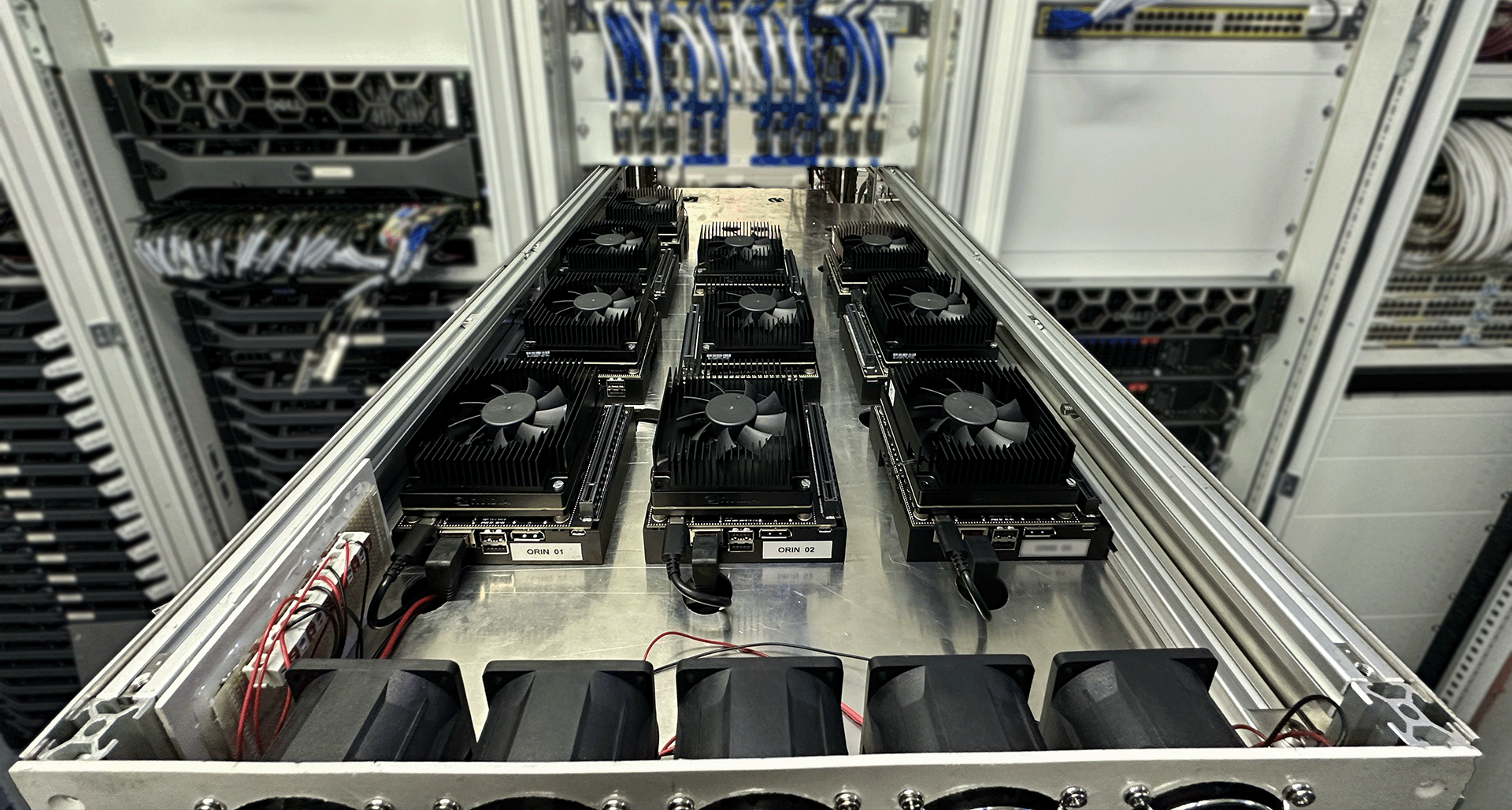}
    \caption{Our NVIDIA Jetson AGX Orin 64GB Testbed. 10 devices with freely configurable network interconnect up to 10 Gbit. Active external cooling is a must at the given energy density of $10 \cdot 60W$ max. power draw.}
    \label{fig:testbed}
    \vspace{-16pt}
\end{figure}

\noindent\textbf{Evaluation hardware}. In our hardware-centered study, we focus on state-of-the-art deep learning accelerators for embedded and data center computing. We employ a cloud VM with a single NVIDIA A100 80~GB (SXM4) as a data center node (A100) to perform our local baseline experiments. 
Further, we use a dedicated cluster consisting of ten NVIDIA Jetson AGX Orin 64~GB nodes (Orin) as the only state-of-the-art embedded computing platform that provides enough computational resources for training FMs. The Orins are connected with a 1~Gbit synchronous network link and are monitored with 2~Hz for their power metrics (\Cref{fig:testbed}). 
For our FL experiments, we use a GPU-accelerated VM co-located with the Orins to handle the model aggregation and testing of the global model. For all of our experiments, we do not limit hardware capabilities.

\noindent\textbf{DL models}. 
For our experiments, we adopt the FLAN-T5 transformer model family \cite{Chung2022} for conditional text generation. 
Even though the FLAN-T5 models' parameter sizes are small compared to other state-of-the-art FMs, they often provide the best-in-class performance \cite{Fu2024}. 
We evaluate the computational training performance of the FLAN-T5-Small model with 80M parameters or 308~MB in size, the FLAN-T5-Base model with 250M parameters (990~MB), the FLAN-T5-Large model with 783M parameters (3.1~GB), and the FLAN-T5-XL model with 3B parameters (11.4~GB). 
For all models, we use their corresponding pre-trained tokenizers. 
For each model, we apply parameter-efficient fine-tuning (PEFT) in the form of Low-Rank Adaptation (LoRA), which is used to reduce the number of trainable parameters to $< 1\%$ of all model parameters \cite{Hu2021}.
We parameterize LoRA for the FLAN-T5 model family as follows: $r = 16$, $\alpha_l = 32$, $\mathrm{dropout} = 0.05$. We do not fine-tune the LoRA bias.

\noindent\textbf{Dataset}. 
All the FLAN-T5 models are fine-tuned on the Samsum dataset with the objective of summarizing texts with a maximum token length of 512 elements \cite{Gliwa2019}. 
The maximum model output length is 95 tokens, which can be translated into the summaries of the respective inputs. 
For our FL experiments, we choose to sample to the number of samples per client subset from a Dirichlet distribution as it is frequently used in related work~\cite{fedscale-icml22, He2020_fedml, caldas2018}. 

\begin{figure*}[!ht]
    \begin{subfigure}[!ht]{0.38\textwidth}
        \raisebox{-\height}{\includegraphics[width=\textwidth]{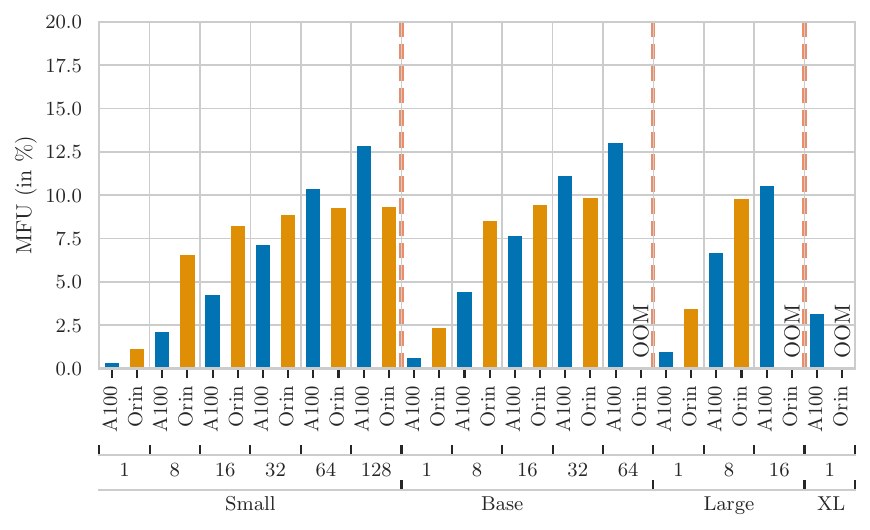}}
        \caption{MFU in \%}
        \label{fig:mfu}
    \end{subfigure}
    \begin{subfigure}[!ht]{0.38\textwidth}
        \raisebox{-\height}{\includegraphics[width=\textwidth]{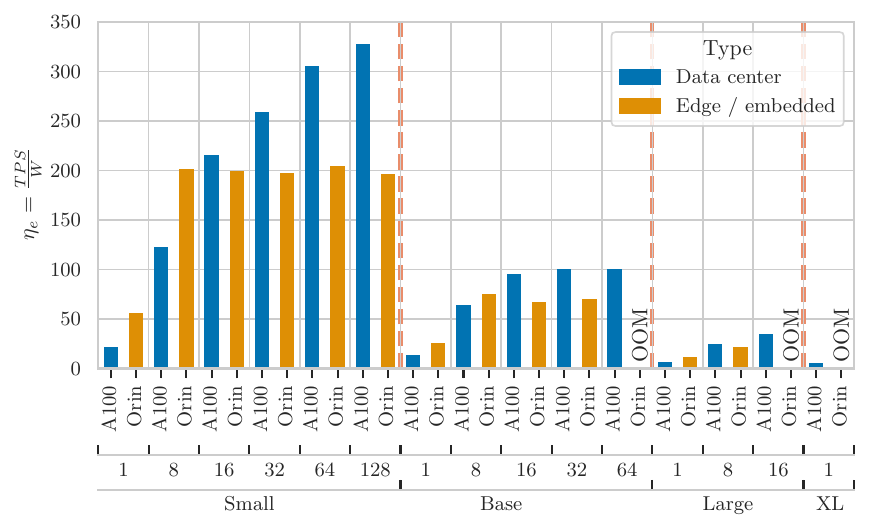}}
        \caption{$\eta_e$ in $\frac{\mathrm{TPS}}{W}$}
        \label{fig:energy}
    \end{subfigure}
    \begin{subfigure}[!ht]{0.18\textwidth}
        \raisebox{-\height}{\includegraphics[width=\textwidth]{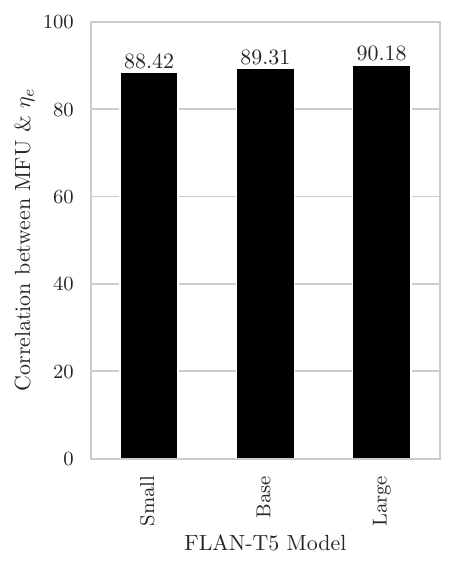}}
        \caption{Pearson Correlation}
        \label{fig:ee_mfu_cor}
    \end{subfigure}
        
    \caption{We study the model FLOP utilization (MFU) and the energy efficiency ($\eta_e$) of the FLAN-T5 transformer model family and find a strong correlation between the MFU and $\eta_e$, which is useful to evaluate root causes for poor training speeds in real-time.}
    \label{fig:mfu-energy-comparison}
\end{figure*}

\noindent\textbf{FL setup}. 
We use a Dirichlet $\alpha_d = 1$ to randomly split the Samsum dataset into 100 subsets that we distribute on the Orin compute cluster.
We train all FLAN-T5 models until they overfit the Samsum dataset or have seen $60,000$ samples. 
In each FL round, we let 10 physical clients participate, i.e., we have a participation rate of 10\%.
For each round, we perform 2 local training steps on each client before communicating with the server.
The client-side optimizer is SGD with a learning rate of $1.0$ and no momentum in all experiments. 
On the server side, we employ FedAvg, FedAvgM, FedAdam, and FedAdamW. 
The exact hyperparameters for the server-side optimizer can be drawn from \Cref{tab:hparams} in \Cref{app:exp-setup}.

%% file: chapters/04_results.tex
Our results are organized so that we first understand the computational limitations of what foundation model sizes can be deployed at the edge. 
We analyze to what point scaling primitives, as we know them from data center environments, hold true at the network edge. 
We then study computational bottlenecks and show the limitations of theoretical analysis. 
Next, we investigate the energy efficiency of training with varying minibatch sizes and point out how energy efficiency can be used to estimate the most energy-efficient on-device training configuration.
We round off our systems analysis with a communication cost estimation. 
Lastly, we study the importance of decoupled weight decay when training state-of-the-art foundation models and quantify the cost savings.

\vspace{-6pt}
\subsection{Computational \& Energy Efficiency}
When designing FL systems, we have to anticipate what type of clients will be participating and how we can optimally use their training performance to receive a trained model in a timely manner.
This is especially relevant for edge computing environments where client hardware is significantly distinct from data center hardware typically used to pre-train and prepare foundation models before using them in a federated setting \cite{zhang-etal-2023-fedpetuning, Babakniya2023}. 

\noindent\textbf{Increasing the minibatch size on embedded devices does not scale well}. 
To understand local training performance in detail, we measure the timing using a microbenchmark (\Cref{fig:microbench}). 
We find linearly growing \texttt{opt.step()} times for all models as we scale the minibatch size on the Orins, while the step times on the A100 platform scale logarithmically with increasing batch size (\Cref{fig:microbench}).
The same is true for the backward step.

\noindent\textbf{The Orin platform is severely bottlenecked by memory bandwidth compared to the A100}.
To develop an understanding of what needs to be done to move from linear to logarithmically growing training times, we look at the MFU (\Cref{fig:mfu}).
The MFU can explain whether training exhibits a computational or memory bottleneck on an FL client.
Throughout all experiments, we find a stagnating MFU for the Orins as we scale the minibatch size, while on the A100 the MFU steadily grows.
As such, on the embedded platform, we reach the maximum theoretical computational efficiency with a minibatch size of 64 for FLAN-T5 Small, 32 for FLAN-T5 Base, and 8 for FLAN-T5 Large. 
Overall, a stagnating MFU as minibatch sizes increase means that increased parallel computation potential does not result in additional used FLOPs.
This can only happen if we encounter a memory bandwidth bottleneck.
We see from \Cref{fig:microbench} that the Orin \texttt{opt.step()} function updating model weights and biases is taking up a significant amount of time in comparison to A100, which suggests that its performance is highly dependent on memory bandwidth.

\noindent\textbf{With our proposed energy efficiency metric $\eta_e$, we enable real-time monitoring of computational efficiency on the client-level}.
We study $\eta_e$ and MFU of the NVIDIA A100 and Jetson AGX Orin across the FLAN-T5 transformer family. For the FLAN-T5- Small model, as we scale the batch size, we notice an increasing $\eta_e$ until a minibatch size of 8 (\Cref{fig:energy}). Afterwards, $\eta_e$ remains constant, i.e., scaling the minibatch size further does not yield any performance benefits. The A100, for the same set of experiments, consistently scales with increasing minibatch size.
The evaluation of the MFU on the same set of experiments as for $\eta_e$ unveils an identical trend (\Cref{fig:mfu}).
The correlation between the MFU and $\eta_e$ originates from both metrics being tied to power draw via FLOPs and Tokens per Second (TPS), respectively.

\noindent\textbf{We reach the computational limits of state-of-the-art deep learning accelerators much earlier than theoretical analysis indicates}.
While the MFU suggests we should scale the minibatch size on the Orins up to 128 samples, we find that the actual memory bottlenecks appear at much smaller minibatch sizes already.
When evaluating the energy efficiency during training, we found that we had already reached the highest efficiency on the Orins with a smaller minibatch size of 8 for all models compared to the A100.
As such, we identified the practical computational limits of state-of-the-art embedded devices for DL. 
% As such, when aiming to maximize energy efficiency across the entire system involving state-of-the-art embedded hardware, we have limited options for on-device optimization. 

% For the A100 the same trend is observable on the FLAN-T5 base model.

\begin{figure}
    \centering
    \begin{subfigure}[!ht]{0.47\textwidth}
        \raisebox{-\height}{\includegraphics[width=\textwidth]{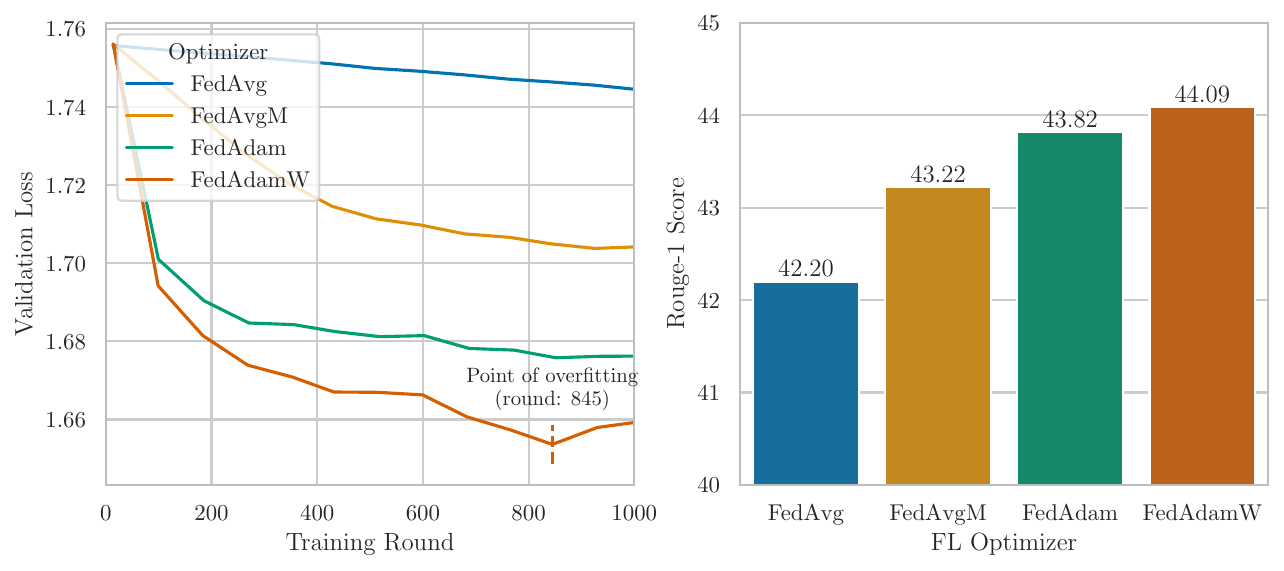}}
        \caption{FLAN-T5 Small}
        \label{fig:flan-t5-small-perf}
    \end{subfigure}\vfill
    \begin{subfigure}[!ht]{0.47\textwidth}
        \raisebox{-\height}{\includegraphics[width=\textwidth]{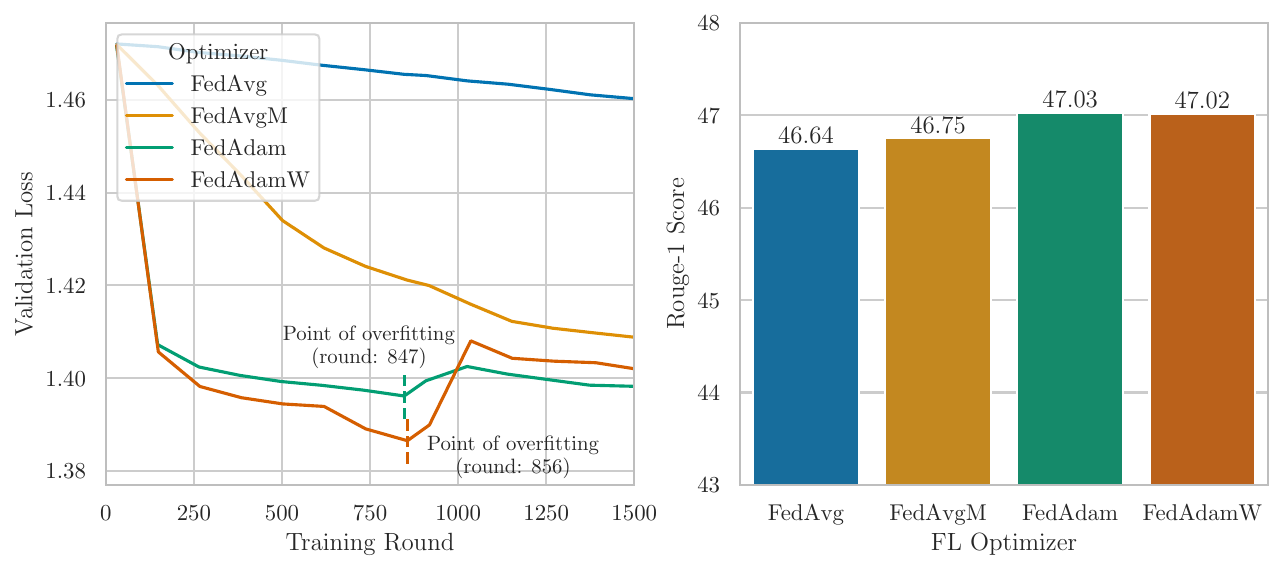}}
        \caption{FLAN-T5 Base}
        \label{fig:flan-t5-base-perf}
    \end{subfigure}\vfill
    \begin{subfigure}[!ht]{0.47\textwidth}
        \raisebox{-\height}{\includegraphics[width=\textwidth]{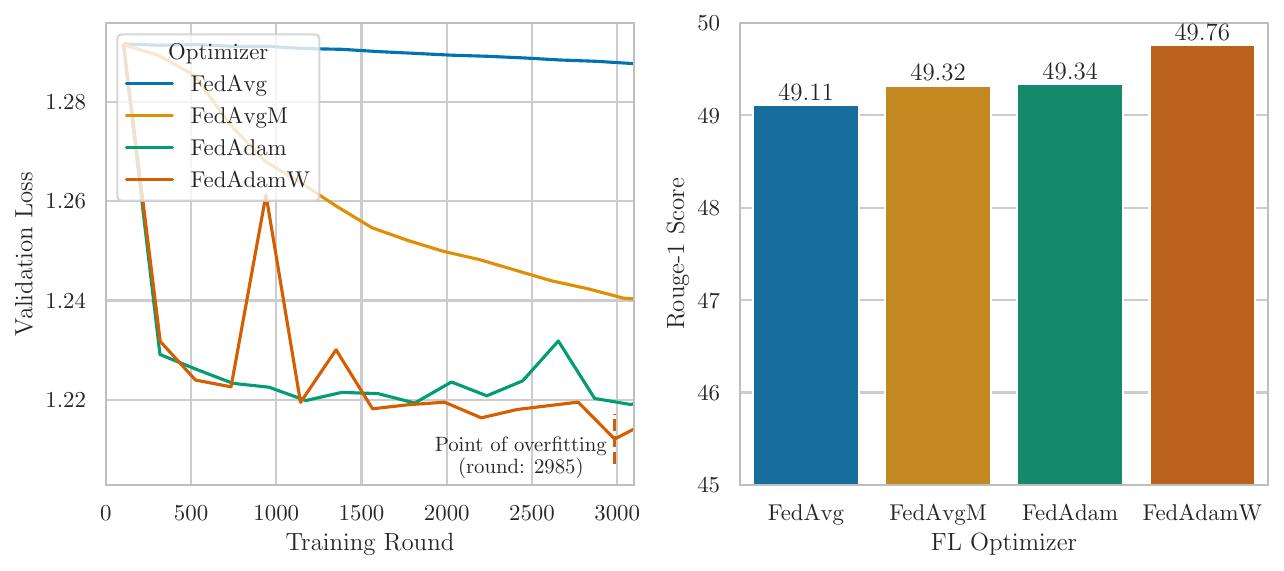}}
        \caption{FLAN-T5 Large}
        \label{fig:flan-t5-large-perf}
    \end{subfigure}\vfill

    \caption{We show the effectiveness of Federated AdamW by training the FLAN-T5 Model family in a federated setup with 100 clients (10 clients per round). We report the validation loss (left) and the Rouge-1 score (right) as performance indicators. Note: The loss spikes for FLAN-T5 Large originate from an increased sensitivity of LoRA adapters with a large parameter count to non-IID data \cite{Babakniya2023}.}
    \label{fig:flan-t5-perf}
    \vspace{-12pt}
\end{figure}

\subsection{Model Performance}
Equally important to the hardware performance side is evaluating state-of-the-art FL optimizers on the algorithmic side, as this affects the energy efficiency of the entire FL system as well.
We find adaptive optimization to be vital for FMs in FL applications.

\noindent\textbf{Federating AdamW accelerates model convergence and saves cost and time compared to other widely-used FL optimizers}.
We compare four commonly used FL optimizers to find out about their training efficiency and convergence speed with state-of-the-art foundation models (\Cref{fig:flan-t5-perf}).
While the FLAN-T5 model family exhibits a slow convergence speed with FedAvg that is approximately three orders of magnitude slower than FedAdam or FedAdamW, we find notable training progress with the adaptive optimizers after 135, 150, and 340 rounds for FLAN-T5 Small, Base, and Large, respectively.
Also, for FLAN-T5 Base and Large, the achievable loss with FedAdamW is lower than with FedAdam before the model starts to overfit the Samsum dataset. 
As such, applying the state of the art for optimization in an FL application yields not only time and cost benefits but also improves model quality at the same time. 

\begin{table}
    \caption{Communication cost analysis when training the FLAN-T5 model family in an FL system with FedAdamW until the minimal loss is achieved. $G \gg 1$ suggests that a model is well-suited for FL at scale. kWh denotes the power consumption incurred during communication per FL round based on the per-bit communication model.}
    \label{tab:granularity_eval}
    \resizebox{0.48\textwidth}{!}{
        \input{tables/networking_granularity}
    }
\end{table}

\subsection{Communication Efficiency}
As we have shown, the computational optimization potential on state-of-the-art embedded hardware is limited. 
As such, it is key to consider the cost of communication during FL training and study how well a model can scale under limited communication.

\noindent\textbf{PEFT significantly improves the scalability of FL systems, regardless of whether bandwidth-limited wireless communication is involved}.
During our experiments, we find PEFT improves $G$ by up to $110\times$ as compared to full model training (\Cref{tab:granularity_eval}). 
This originates from a compounding effect that is beneficial in FL setups. 
Not only does PEFT reduce the demand for computational resources (esp. GPU memory), but by reducing the number of trainable parameters to $< 1\%$ of the total parameter count, it also reduces communication by $> 99\%$.
Due to the relatively higher timeshare of computation compared to communication, this significantly increases $G$, indicating better scalability of an FL application regardless of the communication technology.
At the same time and as expected, full model fine-tuning is only viable in environments that benefit from a high network bandwidth, often absent in FL.

%% file: tables/networking_granularity.tex
\begin{tabular}{lll|rrrrrr}
    \toprule
                &               &                       & \multicolumn{2}{c}{\textbf{FLAN-T5 Small}}    & \multicolumn{2}{c}{\textbf{FLAN-T5 Base}}     & \multicolumn{2}{c}{\textbf{FLAN-T5 Large}} \\
                &               &                       & \multicolumn{2}{c}{(845 FL rounds)}          & \multicolumn{2}{c}{(856 FL rounds)}           & \multicolumn{2}{c}{(2985 FL rounds)} \\
    \textbf{Training} & \textbf{Comm.} & \textbf{Device} & \textbf{$G$} & kWh       & \textbf{$G$} & kWh  & \textbf{$G$} & kWh \\
    \midrule
    \multirow{4}{*}{Full Model} & \multirow{2}{*}{LTE} & A100       & 0.01   & \multirow{2}{*}{0.81}    & 0.00 & \multirow{2}{*}{2.60} & 0.0 & \multirow{2}{*}{8.22} \\  
                                &                      & Orin       & 0.03   &                           & 0.01 &                         & 0.0 &                         \\
                                & \multirow{2}{*}{1 Gbit} & A100    & 14.90  & \multirow{2}{*}{0.13}    & 4.64 & \multirow{2}{*}{0.40}   & 1.47 &  \multirow{2}{*}{1.27}    \\
                                &                      & Orin       & 32.90  &                           & 10.23 &                         & 3.24 &                         \\
    \midrule
    \multirow{4}{*}{PEFT}       & \multirow{2}{*}{LTE} & A100       & 1.70    & \multirow{2}{*}{0.01}     & 0.65 & \multirow{2}{*}{0.02}    & 0.25 & \multirow{2}{*}{0.05} \\
                                &                      & Orin       & 3.70    &                          & 1.44 &                          & 0.54 & \\
                                & \multirow{2}{*}{1 Gbit} & A100    & 1690.90 & \multirow{2}{*}{$< 0.01$}    & 664.29 & \multirow{2}{*}{$< 0.01$}  & 244.74 & \multirow{2}{*}{0.01}  \\
                                &                      & Orin       & 3727.30 &                          & 1464.29 &                      & 539.47 & \\
    \bottomrule
    \bottomrule
\end{tabular}

% LTE Upload speed:             15.69 Mbit/s
% LTE Download speed:           40.39 MBit/s
% 1G Up-/Download speed:        1000 MBit/s

%% file: chapters/05_related_work.tex
We divide our related work section into two major streams of work. 
One is foundation model training with FL, and the other is energy-aware or energy-efficient FL.

\noindent\textbf{Foundation model training with FL}. 
With FATE-LLM, \citet{Fan2023} present an extension of the FATE FL framework to train foundation models, specifically large language models, in a federated setting. 
They introduce a broad range of foundation models and parameter-efficient training techniques with a brief evaluation of the communication benefits of parameter-efficient fine-tuning techniques by means of trainable parameters.
Similarly, FedML \cite{He2020_fedml} supports the training of foundation models in FL systems by providing a wide range of models ready to use.
At the same time, we see a wide variety of parameter-efficient FL methods emerge that tackle data heterogeneity and address resource limitations.
SLoRA~\cite{Babakniya2023} presents a method to tackle the challenge of non-IID data by calibrating the LoRA parameterization in a warm-up phase over multiple rounds, achieving stronger model performance than LoRA without calibration. 
FwdLLM~\cite{Xu2023} enables backpropagation-free fine-tuning of foundation models with a special focus on reducing the memory footprint, enabling the training of models on resource-constrained clients.

\noindent\textbf{Energy-aware and energy-efficient FL}. 
Energy-aware FL system design has been discussed extensively, especially in the space of edge computing applications \cite{shinde2021design,zheng2021federated,yu2021toward,Ye2020,wang2019edge}.
The objective is to reduce the communication cost to a minimum while not compromising model quality.
Yet, most works neglect the full cost of communication as it was introduced by the per-bit communication model \cite{Jalali2014}. 
\citet{Yousefpour2023} provide a holistic viewpoint on energy consumption of a wide range of Fl system configurations. 
Especially, asynchronous FL, which accelerates the training process based on increased training parallelism, incurs a significantly higher energy footprint than round-based FL.

\noindent To the best of our knowledge, there is no overlap between federated foundation model training and energy-aware FL yet. 
Our paper creates this link by evaluating state-of-the-art hardware for its capabilities to serve FL workloads involving foundation models and discusses what can be done to improve the overall energy efficiency of an FL system.
Our evaluation underpins the importance of developing parameter-efficient training techniques for FL, not only to mitigate data heterogeneity effects but also to reduce energy consumption and improve computational efficiency.

%% file: chapters/06_discussion.tex
With formal regulations for AI applications on the horizon, energy monitoring and building energy-efficient FL systems will soon become a necessity to comply with standards for modern AI systems and subsequently build trust with end users.
Therefore, it is key to understand the benefits of FL when working with foundation models (the good), the open challenges (the bad), and what indirect effects FL has (the ugly). 

\noindent\textbf{The Good}. 
By design, FL enables data parallel training of a shared model across geo-distributed clients. 
A major benefit is the access to a much broader range of data as compared to centralized learning where training data is challenging to acquire.
At the same time, the privacy-preserving design of FL also supports building the trust of end users in FL applications since clients must not share their raw data. 
Overall, this helps improve the quality of foundation models on downstream tasks and lower the data bias as we continuously train over an evolving client basis.

\noindent\textbf{The Bad}.
We do find state-of-the-art embedded devices for deep learning applications to be bottlenecked, which limits the applicability of optimization techniques that we know from deep learning in high-performance computing environments, especially as in data centers, the memory bandwidth of GPUs has increased significantly (e.g., with the NVIDIA H100). 
However, as we show in the introduction, the trend of increasing memory bandwidth has also started for embedded devices.
Nonetheless, we can develop targeted optimizations for FL workloads on embedded devices such as the Orins by profiling what GPU kernels are responsible for the on-client memory bottleneck.
Also, promising techniques such as 1.58-bit training of foundation models are capable of reducing the need for high memory bandwidth significantly \cite{Ma2024}.
Furthermore, recent research has shown that LoRA is more sensitive towards a non-IID data distribution, but adaptive methods for configuring LoRA are a promising direction to mitigate this challenge~\cite{Babakniya2023}. 

\noindent\textbf{The Ugly}.
We show in our study that even though we apply PEFT for all FLAN-T5 models, the energy consumption incurred during training and attributable to communication is still significant. To put the energy consumption into perspective: 
Fine-tuning FLAN-T5 Large over the Samsum dataset is possible on a single GPU or even on a single Orin, neglecting the benefit of broader data access, which FL provides. 
With our configuration (\Cref{tab:hparams}), training on an A100 takes approximately 3.33 hours or 1.3 kWh of power, and on an Orin, it takes approximately 8.33 hours or 0.5 kWh.
As such, fine-tuning FLAN-T5 Large consumes more energy for communicating model updates than for the computations. 
This points out the need for future research on even more communication-efficient FL methods than we currently have available. A promising step is gradient projection based on probability-differentiated seeds \cite{Qin2023}.

%% file: chapters/07_conclusion.tex
In our work, we conduct an end-to-end study for FL workloads focusing on energy consumption involving three foundation models.
We point out the hardware limits of state-of-the-art embedded hardware for deep learning and put the performance into perspective with modern data center hardware to discover distinct performance characteristics.
We further introduce $\eta_e$, the real-time metric to evaluate computational bottlenecks, as a drop-in replacement for MFU and show significant potential for on-client optimizations.
Additionally, we show the effectiveness of $\eta_e$ as a proxy for MFU.

To understand the impact FL optimizers have on overall energy consumption, we study three widely used FL optimizers and compare them with FedAdamW, which not only improves model convergence speed but also achieves higher model quality.
Based on the FedAdamW experiments, we quantified the trade-off between communication and computation in FL systems with granularity. 

This underpins the relevance of parameter-efficient training techniques to improve communication efficiency in FL systems and render foundation model training practical.
In the course of our communication analysis, we also quantify the end-to-end energy consumption for communication in our FL experiments, showing that communication is orders of magnitude more energy-intensive than computation for an FL application.  
Putting the current state of FL research into context with emerging AI regulation, we find significant benefits of FL over centralized learning when it comes to data lineage and the potential for data bias mitigation but we have to pick up on research for energy-efficient FL system designs. 

To conclude, we demonstrate the feasibility of fine-tuning foundation models in FL systems but we also hope to raise awareness of the substantial challenges that need to be overcome to enable foundation model training on a broad basis for systems suffering from limited computational and network resources. 

%% file: chapters/08_appendix.tex
\section*{Appendix}
Our appendix is organized along the main paper structure.
We provide additional details for our methodology, experimental setup, and experiment results.

\section{Methodology}
\label{app:methodology}
\textbf{Communication efficiency}. We use to per-bit communication model proposed by \citet{Jalali2014} to estimate the total communication cost of our FL experiments.
In the following, we provide a detailed explanation of calculating the cost for an FL experiment with the per-bit communication cost model.

\begin{equation}
\label{eq:communication_cost}
    \begin{aligned}
        P_{\mathrm{t}} = E_t \cdot \mathcal{B} =~& (n_\mathrm{as} \cdot E_\mathrm{as} + n_\mathrm{LTEE} \cdot E_\mathrm{LTEE} + n_\mathrm{LTEB} \cdot E_\mathrm{LTEB} + E_\mathrm{bng} \\ 
        &+ n_e \cdot E_e + n_c \cdot E_c + n_d \cdot E_d) \cdot \mathcal{B}\mathrm{.}
    \end{aligned}
\end{equation}

$P_{\mathrm{t}}$ is the total power draw for a single transmission. 
$E_\mathrm{as}$, $E_\mathrm{LTEE}$, $E_\mathrm{LTEB}$, $E_\mathrm{bng}$, $E_e$, $E_c$, $E_d$ denote the per-bit energy consumption of edge one or more ethernet switches $n_\mathrm{as}$, one or $0$ client-side LTE endpoint $n_\mathrm{LTEE}$, one or $0$ LTE base stations $n_\mathrm{LTEB}$, the broadband network gateway (BNG), one or more edge routers $n_e$, one or more core routers $n_c$, and one or more data center Ethernet switches $n_d$, respectively. 
We adopt the energy consumption of networking devices as specified in \citet{Jalali2014} and assume 
$n_\mathrm{as} = 1$, $n_\mathrm{bng} = 1$, $n_\mathrm{e} = 3$. $n_\mathrm{c} = 4$, and $n_\mathrm{d} = 2$. 
For calculation involving wireless communication, we assume $n_\mathrm{LTEE} = 1$ and $n_\mathrm{LTEB} = 1$, $0$ otherwise.
$\mathbb{B}$ is the number of trainable model parameters multiplied by the parameter precision (32 bits in our case).
\begin{table}[!ht]
    \centering
    \caption{Energy efficiency is measured in $\frac{\mathrm{TPS}}{W}$. }
    \label{tab:energy-efficiency}
    \resizebox{0.4\textwidth}{!}{
        \begin{adjustbox}{angle=0}
            \input{tables/energy_efficiency}
        \end{adjustbox}
    }
\end{table}

\begin{table*}
    \centering
    \adjustbox{max width = 0.99\textwidth}{
        \begin{tabular}{l|rrrr|rrrr|rrrr}
            \toprule
            \textbf{Model}               & \multicolumn{4}{|c}{\textbf{Small}}                                       & \multicolumn{4}{|c}{\textbf{Base}}                                        & \multicolumn{4}{|c}{\textbf{Large}} \\
            \textbf{Optimizer}           & \textbf{FedAvg} & \textbf{FedAvgM} & \textbf{FedAdam} & \textbf{FedAdamW} & \textbf{FedAvg} & \textbf{FedAvgM} & \textbf{FedAdam} & \textbf{FedAdamW} & \textbf{FedAvg} & \textbf{FedAvgM} & \textbf{FedAdam} & \textbf{FedAdamW}\\
            \midrule             
            Mini-Batch Size     & \multicolumn{4}{|c}{30}                                                   & \multicolumn{4}{|c}{20}                                                   & \multicolumn{4}{|c}{10}  \\
            Learning Rate       & 0.01      & 0.1               & 0.0005    & 0.0005                        & 0.001     & 0.1               & 0.0005    & 0.0005                        & 0.01     & 0.1               & 0.0005    & 0.0005   \\           
            Weight Decay        & 0.001     & 0.001             & --        & 0.001                         & 0.001     & 0.001             & --        & 0.001                         & 0.001     & 0.001             & --        & 0.001    \\     
            Momentum            & 0.0       & 0.9               & --        & --                            & 0.0       & 0.9               & --        & --                            & 0.0       & 0.9               & --        & --       \\     
            $\beta_1$           & --          &--                   & 0.9       & 0.9                           & --          &--                   & 0.9       & 0.9                           &--           &--                   & 0.9       & 0.9      \\       
            $\beta_2$           & --          & --                  & 0.999     & 0.999                         &--           & --                  & 0.999     & 0.999                         & --          & --                  & 0.999     & 0.999    \\       
            Training rounds     & \multicolumn{4}{|c}{1000}                                                  & \multicolumn{4}{|c}{1500}                                                  & \multicolumn{4}{|c}{1500}                              \\
            Clients p. Round    & \multicolumn{12}{c}{10}                           \\
            \bottomrule
            \bottomrule
        \end{tabular}
    }
    \caption{Hyperparameter settings for all FLAN-T5 models and the corresponding optimizers.}
    \label{tab:hparams}
\end{table*}

\section{Experimental Setup}
\label{app:exp-setup}

\textbf{Evaluation hardware}. 
We feel it is important to provide an estimate for establishing a research cluster with NVIDIA Jetson Orin 64 GB devices.
We purchased the devices in mid-2023 at a unit price tag of roughly EUR 2,400, totaling EUR 24,000 just in compute.
Additionally, we equipped each Orin with a Samsung 980 Pro 1 TB NVMe SSD, which cost us a total of EUR 700.
The necessary networking infrastructure (FS S5860-48XMG-U + cables) with a 10 Gbit/s uplink for each device had a price tag of EUR 4,900.
The enclosure is custom-made from sheet metal and aluminum to fit a standard 19-inch rack. 
The material for the case was around EUR 150 + 5 hours for assembly.
In total, our embedded computing cluster cost us just shy of EUR 30,000.
We are happy to share CAD designs and a full part list with anyone interested.

\textbf{DL models}.
We use the pre-trained FLAN-T5 models provided by Google via the HuggingFace hub.
For each model, we ran a hyperparameter search in a centralized experiment on a single node. In our experiments, we chose the optimal hyperparameter configuration as depicted in \cref{tab:hparams}. Our search space is as follows. The learning rate was selected from a set of values [0.1, 0.001, 0.0006, 0.0005, 0.0003, 0.0001, 0.00001]. Similarly, we selected the weight decay from the following set [0.1, 0.003, 0.001, 0.0009, 0.0001]. The momentum and $\beta_1$ were selected from the set [0.85, 0.9, 0.95]. $\beta_2$ was selected from the set [0.99, 0.995, 0.999, 0.9999].

\begin{figure*}[!ht]
    \includegraphics[width=\textwidth]{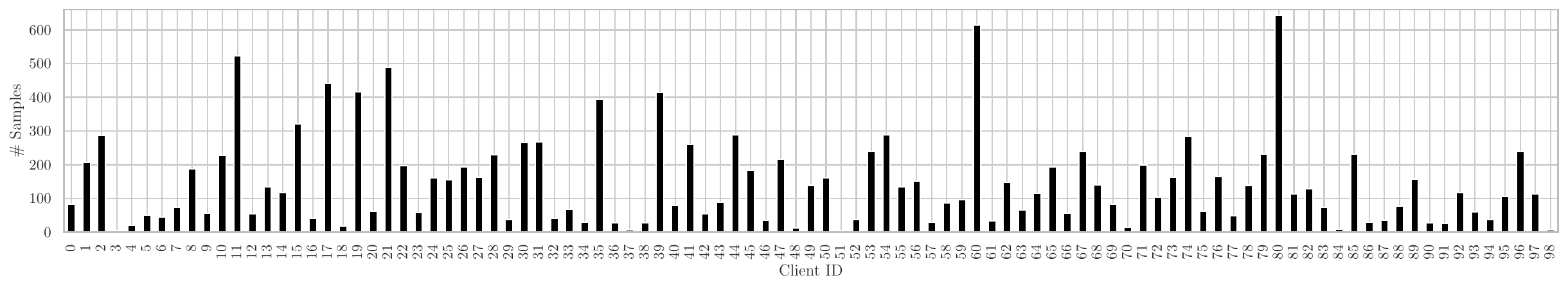}
    \caption{Dataset samples per client}
    \label{fig:sample-dist}
\end{figure*}

\textbf{Dataset}. 
We randomly sample the Samsum dataset by using a Dirichlet distribution ($\alpha = 1.0$) for 100 clients in our experiments. 
The number of samples per client is depicted in \Cref{fig:sample-dist}.

\textbf{FL setup}
In \Cref{tab:hparams}, we provide the full set of hyperparameters used in our experiments.
For our main paper, we limit the number of samples each model sees to $60,000$. 
In addition, we performed additional experiments to identify the point of overfitting for each FL optimizer that would allow a model to converge.
We chose to validate the global model every 200 FL rounds.

\section{Results}
\label{app:results}
This appendix section contains additional results on the energy efficiency measurements and micro-benchmark timings.

\subsection{Energy efficiency}
Energy efficiency is derived from the average power draw of each device during our experiments. The experiments are fixed to 100 steps per epoch for each experiment. \Cref{tab:energy-efficiency} contains details on our energy efficiency calculations.

\subsection{Model FLOP Utilization}
MFU helps to identify computational or memory bottlenecks. \Cref{tab:mfu} depicts all details required to calculate the MFU.

\subsection{Micro-benchmark}
\Cref{tab:microbenchmark-data} describes the step timings in detail and provides a perspective on the speed differences between the data center and embedded hardware.

\begin{table}
    \centering
    \caption{Results of the micro-benchmark of the FLAN-T5 transformer model family on the NVIDIA A100 and Jetson AGX Orin platforms. The sequence length per batch item is 512.}
    \label{tab:microbenchmark-data}
    \resizebox{0.4\textwidth}{!}{
        \begin{adjustbox}{angle=90}
            \input{tables/microbenchmark_flops_table}
        \end{adjustbox}
    }
\end{table}

\begin{table}
    \centering
    \caption{Details on MFU calculation for the NVIDIA A100 and Jetson AGX Orin platforms.}
    \label{tab:mfu}
    \resizebox{0.35\textwidth}{!}{
        \begin{adjustbox}{angle=90}
            \input{tables/mfu}
        \end{adjustbox}
    }
\end{table}

%% file: tables/energy_efficiency.tex
\begin{tabular}{llllll}
\toprule
FLAN-T5 Model & Minib. Size & Device & Avg. Power Draw (W) & $\eta_e$ & TPS \\
\midrule
\multirow[t]{12}{*}{Small} & \multirow[t]{2}{*}{1} & A100 & 75.8 & 21.16 & 1603.62 \\
 &  & AGX Orin & 16.7 & 55.26 & 923.04 \\
\cline{2-6}
 & \multirow[t]{2}{*}{8} & A100 & 103.17 & 121.63 & 12548.15 \\
 &  & AGX Orin & 28.1 & 200.75 & 5641.15 \\
\cline{2-6}
 & \multirow[t]{2}{*}{16} & A100 & 120.96 & 215.03 & 26010.34 \\
 &  & AGX Orin & 35.8 & 198.32 & 7099.35 \\
\cline{2-6}
 & \multirow[t]{2}{*}{32} & A100 & 171.15 & 258.52 & 44246.31 \\
 &  & AGX Orin & 38.84 & 196.53 & 7633.56 \\
\cline{2-6}
 & \multirow[t]{2}{*}{64} & A100 & 212.53 & 304.82 & 64782.11 \\
 &  & AGX Orin & 38.82 & 203.57 & 7903.25 \\
\cline{2-6}
 & \multirow[t]{2}{*}{128} & A100 & 247.72 & 327.2 & 81055.16 \\
 &  & AGX Orin & 41.08 & 195.73 & 8040.93 \\
\cline{1-6} \cline{2-6}
\multirow[t]{10}{*}{Base} & \multirow[t]{2}{*}{1} & A100 & 85.63 & 13.0 & 1113.37 \\
 &  & AGX Orin & 24.86 & 25.23 & 627.45 \\
\cline{2-6}
 & \multirow[t]{2}{*}{8} & A100 & 135.57 & 63.77 & 8645.35 \\
 &  & AGX Orin & 31.3 & 74.21 & 2322.81 \\
\cline{2-6}
 & \multirow[t]{2}{*}{16} & A100 & 159.09 & 94.55 & 15041.23 \\
 &  & AGX Orin & 38.59 & 66.51 & 2566.63 \\
\cline{2-6}
 & \multirow[t]{2}{*}{32} & A100 & 223.55 & 99.28 & 22194.39 \\
 &  & AGX Orin & 38.54 & 69.84 & 2691.45 \\
\cline{2-6}
 & \multirow[t]{2}{*}{64} & A100 & 260.68 & 100.08 & 26088.77 \\
 &  & AGX Orin & \multirow[t]{3}{*}{Out of memory} \\
\cline{1-6} \cline{2-6}
\multirow[t]{6}{*}{Large} & \multirow[t]{2}{*}{1} & A100 & 91.61 & 6.06 & 554.97 \\
 &  & AGX Orin & 26.1 & 11.24 & 293.38 \\
\cline{2-6}
 & \multirow[t]{2}{*}{8} & A100 & 173.43 & 24.24 & 4204.11 \\
 &  & AGX Orin & 41.46 & 20.36 & 844.18 \\
\cline{2-6}
 & \multirow[t]{2}{*}{16} & A100 & 196.9 & 33.76 & 6647.37 \\
 &  & AGX Orin & \multirow[t]{3}{*}{Out of memory} \\
\cline{1-6} \cline{2-6}
\multirow[t]{2}{*}{XL} & \multirow[t]{2}{*}{1} & A100 & 128.21 & 4.31 & 552.0 \\
 &  & AGX Orin & \multirow[t]{3}{*}{Out of memory} \\
\cline{1-6} \cline{2-6}
\bottomrule
\end{tabular}

%% file: tables/microbenchmark_flops_table.tex
\begin{tabular}{lllrrrrrrr}
\toprule
FLAN-T5 Model & Batch Size & Device & Backward & Opt. Step & Loss Calc. & Forward & Batch Loading & TPS & Total Time \\
\midrule
\multirow[t]{12}{*}{Small} & \multirow[t]{2}{*}{1} & A100 & 0.09 & 0.16 & 0.0 & 0.06 & 0.01 & 1657.87 & 0.32 \\
 &  & AGX Orin & 0.15 & 0.3 & 0.0 & 0.09 & 0.01 & 927.51 & 0.55 \\
\cline{2-10}
 & \multirow[t]{2}{*}{8} & A100 & 0.09 & 0.16 & 0.0 & 0.06 & 0.01 & 13051.3 & 0.33 \\
 &  & AGX Orin & 0.22 & 0.4 & 0.0 & 0.1 & 0.01 & 5665.54 & 0.73 \\
\cline{2-10}
 & \multirow[t]{2}{*}{16} & A100 & 0.09 & 0.16 & 0.0 & 0.06 & 0.01 & 26741.79 & 0.31 \\
 &  & AGX Orin & 0.36 & 0.63 & 0.0 & 0.15 & 0.01 & 7112.45 & 1.15 \\
\cline{2-10}
 & \multirow[t]{2}{*}{32} & A100 & 0.12 & 0.18 & 0.0 & 0.06 & 0.01 & 45428.27 & 0.37 \\
 &  & AGX Orin & 0.66 & 1.16 & 0.0 & 0.32 & 0.01 & 7713.02 & 2.15 \\
\cline{2-10}
 & \multirow[t]{2}{*}{64} & A100 & 0.17 & 0.26 & 0.0 & 0.06 & 0.01 & 65944.32 & 0.51 \\
 &  & AGX Orin & 1.28 & 2.22 & 0.0 & 0.64 & 0.01 & 7992.08 & 4.15 \\
\cline{2-10}
 & \multirow[t]{2}{*}{128} & A100 & 0.28 & 0.46 & 0.0 & 0.06 & 0.02 & 82045.0 & 0.81 \\
 &  & AGX Orin & 2.6 & 4.34 & 0.0 & 1.21 & 0.01 & 8046.96 & 8.15 \\
\cline{1-10} \cline{2-10}
\multirow[t]{10}{*}{Base} & \multirow[t]{2}{*}{1} & A100 & 0.14 & 0.23 & 0.0 & 0.08 & 0.01 & 1134.51 & 0.46 \\
 &  & AGX Orin & 0.22 & 0.45 & 0.0 & 0.14 & 0.01 & 631.08 & 0.82 \\
\cline{2-10}
 & \multirow[t]{2}{*}{8} & A100 & 0.14 & 0.23 & 0.0 & 0.09 & 0.01 & 8805.38 & 0.47 \\
 &  & AGX Orin & 0.51 & 0.95 & 0.0 & 0.3 & 0.01 & 2339.49 & 1.76 \\
\cline{2-10}
 & \multirow[t]{2}{*}{16} & A100 & 0.17 & 0.27 & 0.0 & 0.09 & 0.02 & 15380.06 & 0.54 \\
 &  & AGX Orin & 0.93 & 1.69 & 0.0 & 0.57 & 0.01 & 2590.44 & 3.19 \\
\cline{2-10}
 & \multirow[t]{2}{*}{32} & A100 & 0.24 & 0.38 & 0.0 & 0.1 & 0.01 & 22427.21 & 0.74 \\
 &  & AGX Orin & 1.81 & 3.19 & 0.0 & 1.09 & 0.01 & 2692.26 & 6.09 \\
\cline{2-10}
 & \multirow[t]{2}{*}{64} & A100 & 0.4 & 0.65 & 0.0 & 0.19 & 0.01 & 26250.25 & 1.26 \\
 &  & AGX Orin & \multirow[t]{9}{*}{Out of memory} \\
\cline{1-10} \cline{2-10}
\multirow[t]{6}{*}{Large} & \multirow[t]{2}{*}{1} & A100 & 0.27 & 0.46 & 0.0 & 0.18 & 0.02 & 562.2 & 0.92 \\
 &  & AGX Orin & 0.43 & 1.03 & 0.0 & 0.27 & 0.01 & 298.91 & 1.75 \\
\cline{2-10}
 & \multirow[t]{2}{*}{8} & A100 & 0.29 & 0.49 & 0.0 & 0.18 & 0.02 & 4260.38 & 0.97 \\
 &  & AGX Orin & 1.31 & 2.62 & 0.0 & 0.92 & 0.01 & 850.43 & 4.85 \\
\cline{2-10}
 & \multirow[t]{2}{*}{16} & A100 & 0.4 & 0.62 & 0.0 & 0.2 & 0.02 & 6728.79 & 1.23 \\
 &  & AGX Orin & \multirow[t]{9}{*}{Out of memory} \\
\cline{1-10} \cline{2-10}
\multirow[t]{2}{*}{XL} & \multirow[t]{2}{*}{1} & A100 & 0.27 & 0.48 & 0.0 & 0.17 & 0.02 & 560.07 & 0.93 \\
 &  & AGX Orin & \multirow[t]{9}{*}{Out of memory} \\
\cline{1-10} \cline{2-10}
\bottomrule
\end{tabular}

%% file: tables/mfu.tex
\begin{tabular}{lllllllllll}
\toprule
 FLAN-T5 Model & Minib. Size & Device & TPS & Params & \# Layers & $d_{\mathrm{model}}$ & $d_{\mathrm{ff}}$ & $n_{\mathrm{att_heads}}$ & Seq. Len. & MFU \\
\midrule
\multirow[t]{12}{*}{Small} & \multirow[t]{2}{*}{1} & A100 & 1657.0 & 0.0 & 8.0 & 512.0 & 1024.0 & 8.0 & 512.0 & 0.3 \\
 &  & Orin AGX & 927.0 & 0.0 & 8.0 & 512.0 & 1024.0 & 8.0 & 512.0 & 1.1 \\
\cline{2-11}
 & \multirow[t]{2}{*}{8} & A100 & 13051.0 & 0.0 & 8.0 & 512.0 & 1024.0 & 8.0 & 512.0 & 2.0 \\
 &  & Orin AGX & 5665.0 & 0.0 & 8.0 & 512.0 & 1024.0 & 8.0 & 512.0 & 6.5 \\
\cline{2-11}
 & \multirow[t]{2}{*}{16} & A100 & 26741.0 & 0.0 & 8.0 & 512.0 & 1024.0 & 8.0 & 512.0 & 4.2 \\
 &  & Orin AGX & 7112.0 & 0.0 & 8.0 & 512.0 & 1024.0 & 8.0 & 512.0 & 8.1 \\
\cline{2-11}
 & \multirow[t]{2}{*}{32} & A100 & 45428.0 & 0.0 & 8.0 & 512.0 & 1024.0 & 8.0 & 512.0 & 7.1 \\
 &  & Orin AGX & 7713.0 & 0.0 & 8.0 & 512.0 & 1024.0 & 8.0 & 512.0 & 8.8 \\
\cline{2-11}
 & \multirow[t]{2}{*}{64} & A100 & 65944.0 & 0.0 & 8.0 & 512.0 & 1024.0 & 8.0 & 512.0 & 10.3 \\
 &  & Orin AGX & 8040.0 & 0.0 & 8.0 & 512.0 & 1024.0 & 8.0 & 512.0 & 9.2 \\
\cline{2-11}
 & \multirow[t]{2}{*}{128} & A100 & 82045.0 & 0.0 & 8.0 & 512.0 & 1024.0 & 8.0 & 512.0 & 12.8 \\
 &  & Orin AGX & 8094.0 & 0.0 & 8.0 & 512.0 & 1024.0 & 8.0 & 512.0 & 9.3 \\
\cline{1-11} \cline{2-11}
\multirow[t]{10}{*}{Base} & \multirow[t]{2}{*}{1} & A100 & 1134.0 & 0.0 & 12.0 & 768.0 & 2048.0 & 12.0 & 512.0 & 0.6 \\
 &  & Orin AGX & 631.0 & 0.0 & 12.0 & 768.0 & 2048.0 & 12.0 & 512.0 & 2.3 \\
\cline{2-11}
 & \multirow[t]{2}{*}{8} & A100 & 8805.0 & 0.0 & 12.0 & 768.0 & 2048.0 & 12.0 & 512.0 & 4.4 \\
 &  & Orin AGX & 2339.0 & 0.0 & 12.0 & 768.0 & 2048.0 & 12.0 & 512.0 & 8.5 \\
\cline{2-11}
 & \multirow[t]{2}{*}{16} & A100 & 15380.0 & 0.0 & 12.0 & 768.0 & 2048.0 & 12.0 & 512.0 & 7.6 \\
 &  & Orin AGX & 2591.0 & 0.0 & 12.0 & 768.0 & 2048.0 & 12.0 & 512.0 & 9.4 \\
\cline{2-11}
 & \multirow[t]{2}{*}{32} & A100 & 22427.0 & 0.0 & 12.0 & 768.0 & 2048.0 & 12.0 & 512.0 & 11.1 \\
 &  & Orin AGX & 2692.0 & 0.0 & 12.0 & 768.0 & 2048.0 & 12.0 & 512.0 & 9.8 \\
\cline{2-11}
 & \multirow[t]{2}{*}{64} & A100 & 26250.0 & 0.0 & 12.0 & 768.0 & 2048.0 & 12.0 & 512.0 & 13.0 \\
 &  & AGX Orin & \multirow[t]{8}{*}{Out of memory} \\
\cline{1-11} \cline{2-11}
\multirow[t]{6}{*}{Large} & \multirow[t]{2}{*}{1} & A100 & 562.0 & 0.0 & 24.0 & 1024.0 & 2816.0 & 16.0 & 512.0 & 0.9 \\
 &  & Orin AGX & 298.0 & 0.0 & 24.0 & 1024.0 & 2816.0 & 16.0 & 512.0 & 3.4 \\
\cline{2-11}
 & \multirow[t]{2}{*}{8} & A100 & 4260.0 & 0.0 & 24.0 & 1024.0 & 2816.0 & 16.0 & 512.0 & 6.6 \\
 &  & Orin AGX & 853.0 & 0.0 & 24.0 & 1024.0 & 2816.0 & 16.0 & 512.0 & 9.7 \\
\cline{2-11}
 & \multirow[t]{2}{*}{16} & A100 & 6728.0 & 0.0 & 24.0 & 1024.0 & 2816.0 & 16.0 & 512.0 & 10.5 \\
 &  & AGX Orin & \multirow[t]{8}{*}{Out of memory} \\
\cline{1-11} \cline{2-11}
\multirow[t]{2}{*}{XL} & \multirow[t]{2}{*}{1} & A100 & 560.0 & 0.0 & 24.0 & 2048.0 & 5120.0 & 32.0 & 512.0 & 3.1 \\
 &  & AGX Orin & \multirow[t]{8}{*}{Out of memory} \\
\cline{1-11} \cline{2-11}
\bottomrule
\end{tabular}